\documentclass[letterpaper, 10 pt, journal, twoside]{ieeetran}

\usepackage{etoolbox}
\makeatletter
\patchcmd{\@makecaption}
  {\scshape}
  {}
  {}
  {}
\makeatletter
\patchcmd{\@makecaption}
  {\\}
  {.\ }
  {}
  {}
\makeatother
\def\tablename{Table}

\IEEEoverridecommandlockouts                              

\usepackage{times}
\usepackage[ruled,vlined,linesnumbered]{algorithm2e}
\usepackage{multicol}
\usepackage[bookmarks=true]{hyperref}
\usepackage{graphicx}
\usepackage{tabularx}
\usepackage{amssymb}
\usepackage{amsmath}
\usepackage{fancyhdr}
\usepackage{tikz}
\usepackage{bm}
\usepackage{multicol}
\usepackage{listings}
\usepackage{subfigure}
\usepackage{xcolor}
\usepackage{scalerel}
\usepackage[colorinlistoftodos]{todonotes}

\SetAlFnt{\small}
\SetAlCapFnt{\small}
\SetAlCapNameFnt{\small}

\newcommand{\vect}[1]{\mathbf{#1}} 
\newcommand{\vectg}[1]{\boldsymbol{#1}}

\newcommand{\bx}{\mathbf{x}}
\newcommand{\bu}{\mathbf{u}}
\newcommand{\bX}{\boldsymbol{\tau}}
\newcommand{\bU}{\boldsymbol{\xi}}
\newcommand{\bK}{\mathbf{K}}
\newcommand{\bQ}{\mathbf{Q}}

\newcommand{\bnu}{\boldsymbol{\nu}}
\newcommand{\bbeta}{\boldsymbol{\eta}}
\newcommand{\bkappa}{\boldsymbol{\kappa}}
\newcommand{\bpi}{\boldsymbol{\pi}}
\newcommand{\bmu}{\boldsymbol{\mu}}
\newcommand{\bzeta}{\boldsymbol{\zeta}}
\newcommand{\bSigma}{\boldsymbol{\Sigma}}
\newcommand{\bff}{\mathbf{f}}

\DeclareMathOperator*{\argmin}{arg\,min}

\def\NAT@parse{\typeout{IEEEtran error: Attempt to use fake Natbib command 
which is provided to fool Hyperref.}}

\allowdisplaybreaks
\begin{document}

\begin{titlepage}
\vspace*{\fill}
{\large
\copyright 2023 IEEE.  Personal use of this material is permitted. Permission from IEEE must be obtained for all other uses, in any current or future media, including reprinting/republishing this material for advertising or promotional purposes, creating new collective works, for resale or redistribution to servers or lists, or reuse of any copyrighted component of this work in other works. DOI: 10.1109/LRA.2023.3315209}
\vspace*{\fill}
\end{titlepage}

\title{Probably Approximately Correct Nonlinear Model Predictive Control (PAC-NMPC)}
\author{Adam Polevoy$^{1, 2}$, Marin Kobilarov$^{2}$, and Joseph Moore$^{1, 2}$%
\thanks{Manuscript received: May, 7, 2023; Revised August, 1, 2023; Accepted August, 24, 2023.}
\thanks{This paper was recommended for publication by Editor Clement Gosselin upon evaluation of the Associate Editor and Reviewers' comments.
This work was supported by JHU APL internal R\&D} 
\thanks{$^{1}$Adam Polevoy and Joseph Moore are with the Applied Physics Lab, Johns Hopkins University, Laurel, MD 20723 USA
        {\tt\footnotesize adam.polevoy@jhuapl.edu, joseph.moore@jhuapl.edu}}%
\thanks{$^{2} $Marin Kobilarov is with the Whiting School of Engineering, Department of Mechanical Engineering, Johns Hopkins University, Baltimore, MD 21218
        {\tt\footnotesize mkobila1@jhu.edu}}%
\thanks{Digital Object Identifier (DOI): see top of this page.}
}
\maketitle

\begin{abstract}


Approaches for stochastic nonlinear model predictive control (SNMPC) typically make restrictive assumptions about the system dynamics and rely on approximations to characterize the evolution of the underlying uncertainty distributions. For this reason, they are often unable to capture more complex distributions (e.g., non-Gaussian or multi-modal) and cannot provide accurate guarantees of performance. In this paper, we present a sampling-based SNMPC approach that leverages recently derived sample complexity bounds to certify the performance of a feedback policy without making assumptions about the system dynamics or underlying uncertainty distributions. By parallelizing our approach, we are able to demonstrate real-time receding-horizon SNMPC with statistical safety guarantees in simulation and on hardware using a 1/10\textsuperscript{th} scale rally car and a 24-inch wingspan fixed-wing unmanned aerial vehicle (UAV).


\end{abstract}


\IEEEpeerreviewmaketitle

\section{Introduction}
\IEEEPARstart{N}{onlinear} model predictive control (NMPC) has proven to be a powerful approach for controlling high-dimensional, complex robotic systems (e.g., \cite{falanga2018pampc,basescu2020direct,ding2019real,grandia2019feedback}). Nevertheless, although these methods can handle large state spaces, nonlinear dynamics, and system constraints, their performance can be adversely affected by the presence of uncertainty even in the context of real-time replanning \cite{paulson2019efficient}. A number of approaches have been proposed to compensate for this marginal robustness, the simplest of which is to generate a feedback policy to track the current receding-horizon plan (e.g., \cite{basescu2020direct}). These approaches, however, do not account for uncertainty or closed-loop performance during the planning process. To address this, approaches such as robust NMPC (RNMPC) and stochastic NMPC (SNMPC) attempt to reason about the response of the policy to disturbances during planning. This is usually accomplished by making simplifying assumptions about the dynamics and disturbances. For instance, in RNMPC, disturbance sets may be approximated by ellipsoids \cite{manchester2017dirtrel}, or robustness guarantees may require the existence of a control contraction metric (CCM) \cite{singh2017robust}; in SNMPC, uncertainty distributions are often approximated via Gaussians \cite{ozaki2020tube} or sampling \cite{yin2022trajectory}. 

In this paper, we present a sampling-based SNMPC algorithm capable of controlling stochastic dynamical systems without applying limiting assumptions to the structure of the stochastic dynamics or underlying uncertainty distributions. In addition, our approach leverages recently derived sample complexity bounds \cite{kobilarov2015sample} to provide a probabilistic guarantee on system performance. Our algorithm builds on Probably Approximately Correct Robust Policy Search (PROPS) \cite{sheckells2020probably} to directly optimize an upper confidence bound on the expected cost and the probability of constraint violation for a receding-horizon feedback policy. Not only does this approach encourage robust policy generation, but the minimized bounds themselves provide a statistical guarantee for each planning interval. To achieve real-time performance, we use a graphics processing unit (GPU) to parallelize our algorithm. We then evaluate our approach in simulation and on hardware using a 1/10\textsuperscript{th} scale rally car and a 24-inch wingspan fixed-wing UAV. 

Our contributions are:


\begin{enumerate}
    \item A novel algorithm, PAC-NMPC, for receding horizon SNMPC with probabilistic performance guarantees.
    \item A real-time implementation via GPU acceleration.
    \item Demonstration of our algorithm on complex underactuated systems via simulation and hardware experiments.
\end{enumerate}

\vspace{-1pt}

\begin{figure}[]
    \centering
    \includegraphics[trim={0 0 0 0},clip,width=0.85\columnwidth]{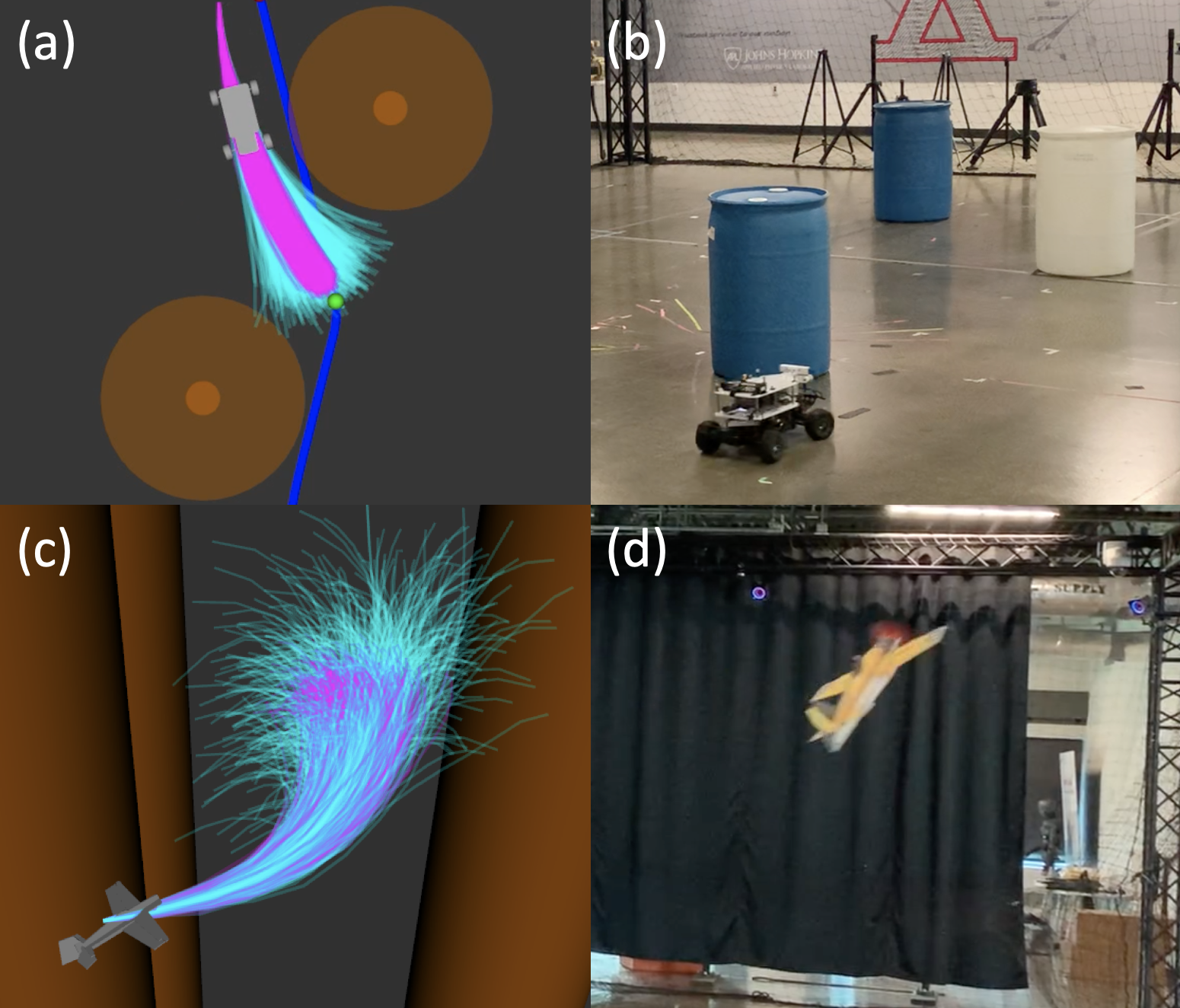}
    \caption{PAC-NMPC optimizes high confidence bounds on the expected costs/constraint of feedback policy distributions. We demonstrated our approach on a 1/10th-scale Rally Car (a, b) and an Edge 540 fixed-wing UAV (c, d).}
    \label{fig:main_img}
\end{figure}

\section{Related Work}

Prior research addresses the impact of uncertainty on NMPC via robust NMPC (RNMPC) or stochastic NMPC (SNMPC). RNMPC represents uncertainty in the system as bounded disturbances. While min-max tube NMPC methods \cite{manchester2017dirtrel, villanueva2017robust, garimella2018robust} and multi-stage NMPC approaches \cite{lucia2014multi} have been proposed, constraint-tightening based tube NMPC \cite{mayne2011tube} has proved to be the most common and computationally tractable RNMPC approach. Most of these approaches rely on tube size derived from a computed Lipschitz constant \cite{marruedo2002input,gao2014tube}, CCMs \cite{singh2017robust, bayer2013discrete,chou2021model}, or bounds on incremental stability \cite{kohler2020computationally}. These methods for computing invariant sets are often complex, may place significant restrictions on the dynamics, and can lead to overly conservative results. Recent work has proposed sampling-based reachability analysis for arbitrary continuously differentiable dynamics with less conservative results \cite{lew2021sampling, wu2023robust}.



Stochastic NMPC represents uncertainty as a probability distribution and optimizes the expectation over costs and constraint violations. Oftentimes, they optimize a feedback policy, rather than a control sequence. Our method, PAC-NMPC, falls into this category of approaches. Early approaches focused on a local iterative solution to the stochastic Hamilton-Jacobi-Bellman equation \cite{todorov2005generalized,todorov2009iterative}, which relies on a first order approximation of the dynamics for noise propagation. Stochastic differential dynamic programming approaches \cite{theodorou2010stochastic} would later allow for second order approximation of the dynamics. More recent methods utilize the unscented transform to propagate noise through nonlinear dynamics \cite{ozaki2020tube, howell2021direct}. In \cite{williams2016aggressive}, the authors leverage path integral control to generate open-loop trajectories and create a sampling-based NMPC algorithm. More recent extensions \cite{yin2022trajectory, williams2018robust} include a feedback policy for robust performance, \cite{gandhi2021robust} provides guarantees on free-energy growth, and \cite{yin2023risk} penalizes Conditional Value-at-Risk (CVaR).


Sampling-based stochastic optimal control methods are closely related to policy search in reinforcement learning. A general background and survey of common policy search methods can be found in \cite{deisenroth2013survey}. In \cite{sheckells2020probably}, researchers developed an approach for robust policy search using a Probably Approximately Correct (PAC) learning framework called PAC Robust Policy Search (PROPS) based on recently derived sample complexity bounds in \cite{kobilarov2015sample}. Researchers have also recently explored (PAC)-Bayes theory to generate collision-avoidance policies with performance guarantees in novel environments \cite{majumdar2021pac} and achieve vision-based planning with motion primitives \cite {veer2021probably}. Our PAC-NMPC method builds on the work in \cite{kobilarov2015sample} and \cite{sheckells2020probably} to develop a highly generalizable sampling-based stochastic feedback motion-planning algorithm that can provide statistical performance guarantees for a large class of uncertain dynamical systems. 

\section{Background}

Before presenting our approach for Probably Approximately Correct NMPC, we first review the recently derived PAC bounds for Iterative Stochastic Policy Optimization which are the foundation for our approach.


\subsection{Iterative Stochastic Policy Optimization}

Consider the stochastic dynamics given by $p(\bx_{t+1}|\bx_{t}, \bu_{t})$ defined by a vector of state values $\bx_{t}\in\mathbb{R}^{N_x}$, a vector of control inputs $\bu_{t}\in\mathbb{R}^{N_u}$, and probability density $p$. Iterative Stochastic Policy Optimization (ISPO) \cite{kobilarov2015sample} formulates the search for a control policy $\bu_{t}= \bpi_t(\bx_t,\bU)$ as a stochastic optimization problem, where $\bU$ is a set of policy parameters. Specifically, ISPO introduces a surrogate distribution $p(\bU|\bnu)$ where policy parameters $\bU$ are dependent on distribution hyper-parameters $\bnu$. This induces a joint distribution
\begin{align}
p(\bX, \bU|\bnu) = p(\bX|\bU)p(\bU|\bnu)
\label{eq:jointdist}
\end{align}
where $ p(\bX|\bU)$ represents the natural stochasticity of the system and is given as $p(\bX|\bU)=p(\bx_0)\prod^T_{t=0}p(\bx_{t+1}|\bx_{t},\bu_t)$, where $\bu_t=\bpi(\bx_t,\bU)$. Here $\bX$ represents the discrete time trajectory sequence $\{\bx_0,\bu_0, \bx_1,\bu_1 ..., \bu_{N_T}, \bx_{N_T+1} \}$, where $N_T$ is the number of timesteps.


In general, the objective of ISPO is to solve the optimization problem 
$\bnu^* = \argmin_{\bnu} \mathcal{J}(\bnu)$ where $\mathcal{J}(\bnu)~=~\mathbb{E}_{\bX, \bU \sim p(\cdot, \cdot|\bnu)}[J(\bX)]$. To do this, ISPO iteratively samples from Eq. \ref{eq:jointdist} and solves the optimization problem $
\widehat{\bnu}_{i+1}^* = \arg \min_{\bnu} \widehat{\mathcal{J}}(\bnu) + \alpha D(\bnu,\widehat{\bnu}_{i}^*)$ for each iteration $i$ until convergence. Here $\widehat{\mathcal{J}}(\bnu)$ is an empirical approximation of the expected cost, $D$ is a distance between distributions, often the $\mathcal{KL}$-divergence, and $\alpha>0$ is hand-tuned weight or Lagrange multiplier \cite{peters2010relative} computed for a constraint on $D(\bnu,\widehat{\bnu}_{i}^*)$. Alg. \ref{alg:ispo} is a general framework for ISPO.






\begin{algorithm}[!t]
    \caption{Episodic ISPO}\label{alg:ispo}
    Inputs: $\widehat{\bnu}_{0}^*$, $i\leftarrow 0$\;
    \While{termination condition not met}{
          \For {$j=1,\dots,M$}{
              Sample Trajectory: $(\bX_j, \bU_j) \sim p(\cdot, \cdot |\widehat{\bnu}_{i}^*)$\;
              Evaluate Cost: $J_j = J(\bX_j)$\;
          }
        $\widehat{\bnu}_{i+1}^* = \arg \min_{\bnu} \widehat{\mathcal{J}}(\bnu) + \alpha D(\bnu,\widehat{\bnu}_{i}^*)$\;
    $i\leftarrow i+1$
    }
    \Return $\widehat{\bnu}^*_i$
\end{algorithm}

\subsection{PAC Bounds for Stochastic Policy Search}
Instead of directly minimizing an empirical approximation of the expected cost, it can be beneficial to minimize \emph{an upper confidence bound} on the expected cost. Not only does such an approach encourage robust policies, but it also can provide guarantees on future performance. In this paper, we directly optimize the PAC bounds derived in \cite{kobilarov2015sample}. While these bounds are derived in \cite{kobilarov2015sample} and discussed in \cite{sheckells2020probably}, we review them here since they are fundamental to our approach.


In \cite{kobilarov2015sample} the presented PAC bound  $\mathcal{J}^+_\alpha(\bnu)$ takes the form
\begin{align}
\mathcal{J}^+_\alpha(\vectg{\nu})\! \triangleq \! \widehat{\mathcal J}_\alpha(\vectg{\nu}) + \alpha d(\vectg{\nu}) + \Phi_{\alpha}(\delta)
\end{align}
where $\widehat{\mathcal J}_\alpha(\vectg{\nu})$ is a robust estimator \cite{catoni2012challenging} of the expected cost, $d(\vectg{\nu})$ is a distance between distributions, and $\Phi_{\alpha}(\delta)$ is a concentration-of-measure term which accounts for discrepancies between the true mean, $\mathcal{J}(\vectg{\nu})$, and the robust estimator, $\widehat{\mathcal J}_\alpha(\vectg{\nu})$. For a particular choice of $\widehat{\mathcal J}_\alpha(\vectg{\nu})$, $d(\vectg{\nu})$, and $\Phi_{\alpha}(\delta)$, \cite{kobilarov2015sample} proves that $\mathcal{J}^+_\alpha(\vectg{\nu})$ bounds the expected cost $\mathcal{J}(\bnu)$ with a probability of $1-\delta$.

To derive an expression for $\widehat{\mathcal J}_\alpha(\vectg{\nu})$, we must estimate 
\begin{align}
\mathcal{J}(\bnu) =\mathbb{E}_{\bX, \bU \sim p(\cdot, \cdot|\bnu_0)}\bigg[J(\bX)\frac{p(\bU|\bnu)}{p(\bU|\bnu_0)}\bigg].
\end{align}
Because the likelihood ratio $\frac{p(\bU|\bnu)}{p(\bU|\bnu_0)}$ can be unbounded, a robust estimation technique is needed. Following \cite{catoni2012challenging}, the expected value of a random variable $X$ can be approximated as $\mathbb{E}[X] \approx \frac{1}{\alpha M}\sum_{i=0}^M  \psi(\alpha X_i)$ for some and $\alpha>0$, where $X_i$ is a sampled value of $X$ and $\psi(x) = \log\left(1+x+\frac{1}{2}x^2\right)$. Given $L$ prior surrogate distributions with hyper-parameters $\bnu_0$, $\bnu_1$, ... $\bnu_{L-1}$ and $M$ iid samples from each joint distribution, $(\bX_{i0}, \bU_{i0})$, $(\bX_{i1}, \bU_{i1})$, ..., $(\bX_{iM}, \bU_{iM})$ $\sim$ $p(\cdot, \cdot | \bnu_i)$, the robust estimate of the expected cost can then be written as
\begin{align}
\widehat {\mathcal J}_\alpha(\bnu) &\triangleq \frac{1}{\alpha LM} \sum_{i=0}^{L-1} \sum_{j=1}^{M} \psi\left(\alpha \ell_{ij} \right)
\label{eq:robust_estimator} \\
\ell_{ij} &= J(\bX_{ij})\frac{p(\bU_{ij}|\bnu)}{p(\bU_{ij}|\bnu_i)}
\label{eq:likelihood_ratio}
\end{align}


To compute the distance between the distributions, the Renyi divergence, $D_2$, is used. $d(\vectg{\nu})$ is given as
\begin{align}
  d(\bnu)\! &\triangleq \! \frac{1}{2L}\sum_{i=0}^{L-1}b_i^2e^{D_2\left(p(\cdot | \bnu)||(p(\cdot | \bnu_i)\right)} \\
  0 &\leq J(\bX_{ij}) \leq b_i \; \forall j = 0, ..., M \label{eq:divergence_metric}
\end{align}

The concentration-of-measure term $\Phi_{\alpha}(\delta)$ is given as 
\begin{align}
  \Phi_{\alpha}(\delta)\! &\triangleq \frac{1}{\alpha LM}\log\frac{1}{\delta}.
\end{align}
It tightens the bound as the number of samples increases and as the bound confidence, $1-\delta$, decreases.

In addition to cost-functions, we can also incorporate the probability of constraint violation, $\mathcal{C}(\bnu)$, of the form $\mathcal{C}(\bnu)=\mathbb{P}(g(\bX)>0)= \mathbb{E}_{\bX, \bU \sim p(\cdot, \cdot|\bnu)}\left[\mathbb{I}_{\{g(\bX)>0\}}\right]$, where $\mathbb{I}$ is the indicator function. Similarly, we can derive $\mathcal{C}^+_\alpha(\bnu)$ which upper bounds $\mathcal{C}(\bnu)$ with probability $1-\delta$.

To combine costs and constraints, we optimize a weighted sum of $\mathcal{J}^+_\alpha(\bnu)$ and $\mathcal{C}^+_\alpha(\bnu)$, where $\gamma > 0$ is a  heuristically selected weighting coefficient. We solve 
\begin{align}
    \bnu_{i+1} = \bnu^* = \argmin_{\bnu}\min_{\alpha>0} (\mathcal{J}^+_\alpha(\bnu) + \gamma \mathcal{C}^+_\alpha(\bnu))
    \label{eq:weighted_sum}
\end{align}


using a GPU implementation of L-BFGS-B \cite{fei2014parallel}. When optimizing the bounds, we used self-normalized importance weights to compute $\ell_{ij}$ in Eq. \ref{eq:likelihood_ratio} and analytical gradients, as described in \cite{sheckells2020probably}.
\section{Approach}
To formulate a receding-horizon NMPC algorithm that leverages the PAC bounds in \cite{kobilarov2015sample}, we first address the stochastic trajectory optimization problem. We then extend this approach to enable real-time feedback motion planning.
\subsection{PAC Stochastic Trajectory Optimization}
To formulate stochastic trajectory optimization using PAC bounds, we consider a parameterization of a discrete-time open-loop policy $\bu_t = \bpi_t(\bU)=\bzeta_t$, where
$\bU ~= ~\begin{bmatrix} \bzeta_0^T & \bzeta_1^T& ...&  \bzeta_{N_T}^T\end{bmatrix}^T$, $\bzeta_t \in \mathbb{R}^{N_u}$, and $N_T$ is the time horizon for the trajectory.

We parameterize the surrogate distribution $p(\bU|\bnu)$ as a multivariate Gaussian over this discrete-time control trajectory so that $ \bU \sim \mathcal{N}\left(\bU | \bmu, \bSigma \right)$. For computational efficiency, we assume a diagonal covariance matrix $\bSigma$ so the distribution parameters are given as $\bnu \triangleq \begin{bmatrix} \bmu^T, diag(\bSigma)^T\end{bmatrix}^T$ where $ diag(\bSigma) = \begin{bmatrix} \bbeta_0^T & \bbeta_1^T& ...&  \bbeta_{N_T}^T\end{bmatrix}^T$ and $\bbeta_t \in \mathbb{R}^{N_u}$.
Thus, a control trajectory with an $N_u$-dimensional control input and $N_T$ timesteps, the surrogate distribution is parameterized as a $N_u N_T$-dimensional multivariate Gaussian where  $\bmu \in \mathbb{R}^{N_u N_T}$ and $diag(\bSigma) \in \mathbb{R}^{N_u N_T}$.




\newcommand\mycommfont[1]{\footnotesize\ttfamily\textcolor{blue}{#1}}
\SetCommentSty{mycommfont}
\begin{algorithm}[t!]
  \caption{PAC Stochastic Trajectory Optimization}\label{alg:trajopt}
  Inputs: $\widehat{\bnu}^*_0$, $\bx_0$, $i \gets 0$\;
  \While{termination condition not met}{
    \For(\tcp*[f]{GPU Parallelized}){$j=1,\dots,M$}
    {
      $\bX_{ij}, \bU_{ij} \sim p(\cdot, \cdot | \bnu_i, \bx_0)$\;
      $J_{ij} = J(\bX_{ij})$,
      $c_{ij} = c(\bX_{ij})$\;
    }
    $\widehat{\bnu}^*_{i+1} \gets \argmin_{\bnu}\min_{\alpha>0} (\mathcal{J}^+_\alpha(\bnu) + \gamma \mathcal{C}^+_\alpha(\bnu))$\;
    $i \gets i+1$\;
  }
  \Return $\widehat{\bnu}^*_i$, $\mathcal{J}^+_\alpha(\widehat{\bnu}^*_i)$, $\mathcal{C}^+_\alpha(\widehat{\bnu}^*_i)$
\end{algorithm}
\begin{algorithm}[t!]
    \caption{$\bX \sim p(\cdot | \bU, \bx_0)$} \label{alg:traj_model}
    Inputs: $\bU = [\bu_0^T, ..., \bu_{N_T}^T]^T$, $\bx_0$\;
    \For(\tcp*[f]{Stochastic Rollout}){$t=0,\dots,N_T$}
    {
        $\bx_{t+1} \sim p(\cdot | \bx_t, \bu_t)$\;
    }
    Return $\bX = \left\{ \bx_0,\bu_0, \bx_1,\bu_1, ...,\bu_{N_T},  \bx_{N_T+1}\right\}$
\end{algorithm}

For each optimization iteration, $M$ trajectories are sampled from the joint distribution (Eq. \ref{eq:jointdist}) and evaluated using the augmented cost in Eq. \ref{eq:weighted_sum}. Sampling from the joint distribution is achieved by first sampling policy parameters from the surrogate distribution and then using these policy parameters to sample from the stochastic dynamics as shown in Alg. \ref{alg:traj_model}. The augmented costs of the samples are used to find a new set of policy parameters that minimize the weighted sum of the cost and constraint bounds (Alg. \ref{alg:trajopt}). This process iterates until a termination condition (e.g. maximum time, convergence metric, etc.) is reached. The optimization not only returns the optimized policy parameters, but also the optimized bounds themselves.




An analytical formulation for the surrogate distribution is necessary to compute the robust estimates (Eq. \ref{eq:robust_estimator}) and Renyi divergences (Eq. \ref{eq:divergence_metric}). By contrast, the only requirement on the stochastic dynamics is to permit sampling. This property allows our algorithm to generalize to a large class of complex (potentially ``black-box'') stochastic dynamics models. In practice, the cost and constraint functions can also be stochastic (e.g. $J_{ij}, c_{ij} \sim p(\cdot, \cdot | \bX_{ij})$).



\subsection{PAC Feedback Motion Planning}
While the ability to compute PAC open-loop policies is valuable, ideally, we would like to compute closed-loop policies of the form $\bpi_t(\bx_t,\bU)$. A closed-loop policy should enable us to generate tighter state-space trajectory distributions and improve the ability of our policies to satisfy control objectives. However, designing feedback policies is not straightforward and can be very system specific. 
One fairly general approach might be to select a local feedback control policy of the form $\bu_t = \bK_t\left(\bx_t^d - \bx_t\right) + \bu_t^{d}$, where $\bkappa =\{\bK_0, \bK_1, ..., \bK_{N_T}\}$ is a sequence of time-varying gains, $\bK_t\in\mathbb{R}^{N_u\times N_x}$. $\bx_t^d\in\mathbb{R}^{N_x}$ is a nominal state trajectory and $\bu_t^d\in\mathbb{R}^{N_u}$ is a nominal input trajectory. All of the elements of these sequences are parameters of the feedback policy. Assuming the surrogate distribution is a mulitvariate Gaussian with a diagonal covariance matrix, in this case, $\bU\in \mathbb{R}^{2(N_xN_u+N_x+N_u)N_T}$. Given the size of the decision space, this parameterization of the policy is unlikely to be computationally tractable. 
\begin{algorithm}[t!]
    \caption{$\bX \sim p(\cdot | \bU, \bx_0)$ with Feedback}\label{alg:feedback}
    Inputs: $\bU = [\bu_0, ..., \bu_{N_T}]$, $\bx_0$\;
    $\bx_0^d = \bx_0$\;
    \For(\tcp*[f]{Nominal Rollout}){$k=0,\dots,N_T$}
    {
        $\bx_{t+1}^d = \bx_{t}^d + f(\bx_t^d, \bu_t^d)\Delta{t}$\;
    }
    $\bX^d = \left\{ \bx_0^d,\bu_0^d, \bx_1^d, \bu_1^d ..., \bu_{N_T}^d, \bx_{N_T+1}^d\right\}$\;
    $\bkappa = \text{TVLQR}(\bX^d)$\;
    \For(\tcp*[f]{Feedback Rollout}){$t=0,\dots,N_T$}
    {
        $\bu_t = \bK_t\left(\bx_t - \bx_t^d\right) + \bu_t^d$\;
        $\bx_{t+1} \sim p(\cdot | \bx_t, \bu_t)$\;
    }
    Return $\bX = \left\{ \bx_0,\bu_0, \bx_1,\bu_1, ...,\bu_{N_T},  \bx_{N_T+1}\right\}$
\end{algorithm}
To overcome this challenge, we instead use a local closed-loop policy of the form
\begin{align}
     \bu_t = \bK_t(\bX^d(\bU,\bx_0))\left(\bx_t^d(\bU,\bx_0) - \bx_t\right) + \bu_t^d(\bU).
    \label{eq:feedback_policy}
\end{align}
Here we refer to $\bX^d\triangleq\{\bx_0^d,\bu_0^d, \bx_1^d,\bu_1^d ... , \bu_{N_T}^d, \bx_{N_T+1}^d\}$ as the nominal trajectory, which is computed using a nominal deterministic discrete-time dynamics model, $\bff(\bx_t^d,\bu_t^d)$,
\begin{align}
\bx_{t+1}^d=\bx_t^d+\bff(\bx_t^d,\bu_t^d)\Delta t
\label{eq:deterministic_dynamics}
\end{align}
where $\bu_t^d = \bzeta_t$. To sample from this feedback policy, we must first compute the nominal trajectory, $\bX^d$. However, $\bX^d$ itself is dependent on $\bU$; for this reason, we employ a two-stage sampling process. In the first stage, we sample from the trajectory distribution $p(\bX^d|\bU)$ using the deterministic dynamics model (Eq.  \ref{eq:deterministic_dynamics}) and the surrogate distribution $p(\bU|\bnu)$. For each sample trajectory $\bX^d$, we the compute the time-varying gains $\bK_t(\bX^d)$. In the second stage, we sample from the closed-loop stochastic dynamics using the feedback policy (Eq. \ref{eq:feedback_policy}). This approach, summarized in Alg. \ref{alg:feedback}, certifies the feedback policy by allowing our sampled costs and constraints to capture the impact of our local feedback policy on our stochastic dynamics model.

We use the finite horizon, discrete, time-varying linear quadratic regulator (TVLQR) \cite{underactuated} to compute the feedback gains $\bK_t(\bX^d)$. For each sampled nominal trajectory, the gains are computed by integrating the finite horizon discrete-time Riccati equations backward in time using the nominal dynamics linearized about $\bX^d$. 
\subsection{PAC-NMPC}

 Finally, we propose a receding-horizon NMPC approach based on our PAC feedback motion planning algorithm (Alg. \ref{alg:nmpc}). At each planning interval, with a period of $H$, the trajectory distribution and corresponding feedback policy is optimized given the current state, $\vect{x_0}$, and the prior hyper-parameters, $\widehat{\bnu}^*_0$. Then, the trajectory and feedback gains are trimmed by $\frac{H}{\Delta t}$ to account for time passed since the beginning of the planning interval and executed by the controller. The trim operation is given as $Trim(\bkappa)~=~\left\{\bK_\frac{H}{\Delta t}, ..., \bK_{N_T}\right\}$. Hyper-parameters determining the first $\frac{H}{\Delta t}$ time steps of the control trajectory represent control signals that the system is executing during the optimization and are not modified during the optimization. To execute the policy, we use the maximum likelihood estimate of the policy parameters, which is simply the mean, $\bmu$, of the multivariate Gaussian $p(\bU|\bnu)$. 


\begin{algorithm}[!t]
    \caption{PAC-NMPC}\label{alg:nmpc}
    Input: $\widehat{\bnu}^*_0$\;
    $\bx_0 =$ GetCurrentState()\;

    \While{objective not completed}{
      $\widehat{\bnu}^* =$ Optimize($\widehat{\bnu}^*_0, \bx_0$)\tcp*[l]{Alg. 2}
      $\bu^d =$ MaximumLikelihoodEstimate($\widehat{\bnu}^*$)\;
      \For(\tcp*[f]{Nominal Rollout}){$t=0,\dots,N_T$}
      {
        $\bx_{t+1}^d = \bx_{t}^d + f(\bx_t^d, \bu_t^d)\Delta{t}$\;
      }
      $\bX^d = \left\{ \bx_0^d,\bu_0^d, \bx_1^d, \bu_1^d ..., \bu_{N_T}^d, \bx_{N_T+1}^d\right\}$\;
      $\bkappa = \text{TVLQR}(\bX^d)$\;
      $\bX^d, \bkappa, \widehat{\bnu}^* \gets$ Trim($\bX^d, \bkappa, \widehat{\bnu}^*$)\;
      Execute ($\bX^d,\bkappa$)\;
      $\bx_0 =$ GetCurrentState()\;
      $\widehat{\bnu}^*_0 =$ InitializePrior($\widehat{\bnu}^*$,$\bX^d$,$\bkappa$, $\bx_0$)\tcp*[l]{Alg. 6}
    }
    Return
\end{algorithm}


\begin{algorithm}[!t]
  \caption{Initialize Prior}\label{alg:createprior}
  Input: $\widehat{\bnu}^*$, $\bX^d$, $\bkappa$, $\bx_0$\;
  $[\bmu_0^T, \bmu_1^T,  ..., \bmu_{N_T-\frac{H}{\Delta t}}^T, \bbeta_0^T, \bbeta_1^T, ..., \bbeta_{N_T-\frac{H}{\Delta t}}^T]^T = \widehat{\bnu}^*$\;
  $\bu^d = \bmu$\;
  \For(\tcp*[f]{Feedback Rollout}){$t=0,\dots,N_T-\frac{H}{\Delta t}$}
  {
    $\bu_t = \bK_t\left(\bx_t - \bx_t^d\right) + \bu_t^d$\;
    $\bx_{t+1} = \bx_{t} + f(\bx_t, \bu_t)\Delta{t}$\;
  }
  \For(\tcp*[f]{Extend prior}){$t=N_T-\frac{H}{\Delta t}+1,\dots,N_T$}
  {
    $\bu_t = \mathbf{0}$\;
    $\bbeta_t = max(\bbeta_{T-\frac{H}{\Delta t}}, \bbeta_{min})$\;
  }
  Return $\widehat{\bnu}^*_0 = [\bu_0^T, \bu_1^T,  ..., \bu_{N_T}^T, \bbeta_0^T, \bbeta_1^T, ..., \bbeta_{N_T}^T]^T$
\end{algorithm}



Given that optimal trajectory distributions are expected to be similar between subsequent planning intervals, we use the optimal hyper-parameters from the last planning interval to initialize the prior policy distribution for the next. This initialization is extremely important since it allows the prior to start near an optimal solution, thus reducing the required number of iterations for convergence. After trimming, we apply the feedback policy to generate a modified policy parameter distribution with a mean feasible for the current initial state. Then, we extend the mean with zeros and the variance with its final value. We threshold the prior variance with $\bbeta_{min}$ to prevent it from starting too small, which would inhibit exploration (Alg. \ref{alg:createprior}).
\section{Simulation Experiments}

\subsection{Trajectory Optimization Experiments}

\begin{figure*}[!t]
    \centering
    \includegraphics[trim={15 0 45 5},clip,width=1.0\textwidth]{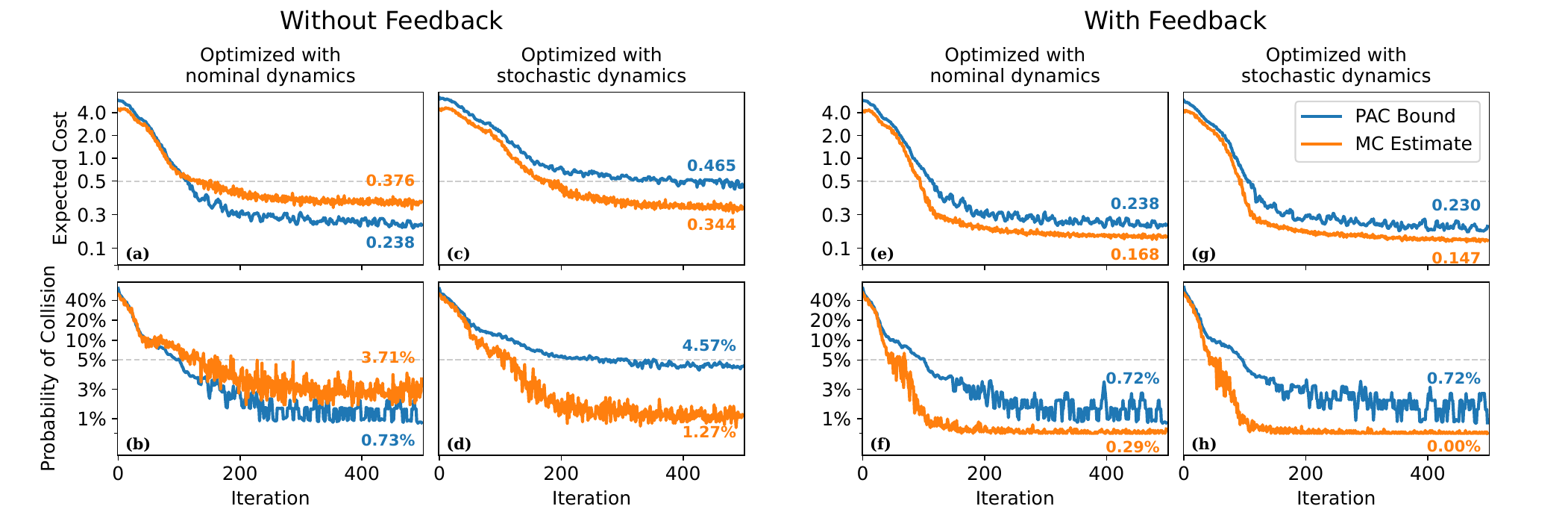}
    \caption{Convergence of PAC Bound compared to MC estimates (evaluated with stochastic dynamics). Y-axis switches from log to linear scale halfway}
    \label{fig:ugv_bounds}
\end{figure*}

 \begin{figure}[t!]
    \centering
    \includegraphics[trim={15 65 45 85},clip,width=1.0\columnwidth]{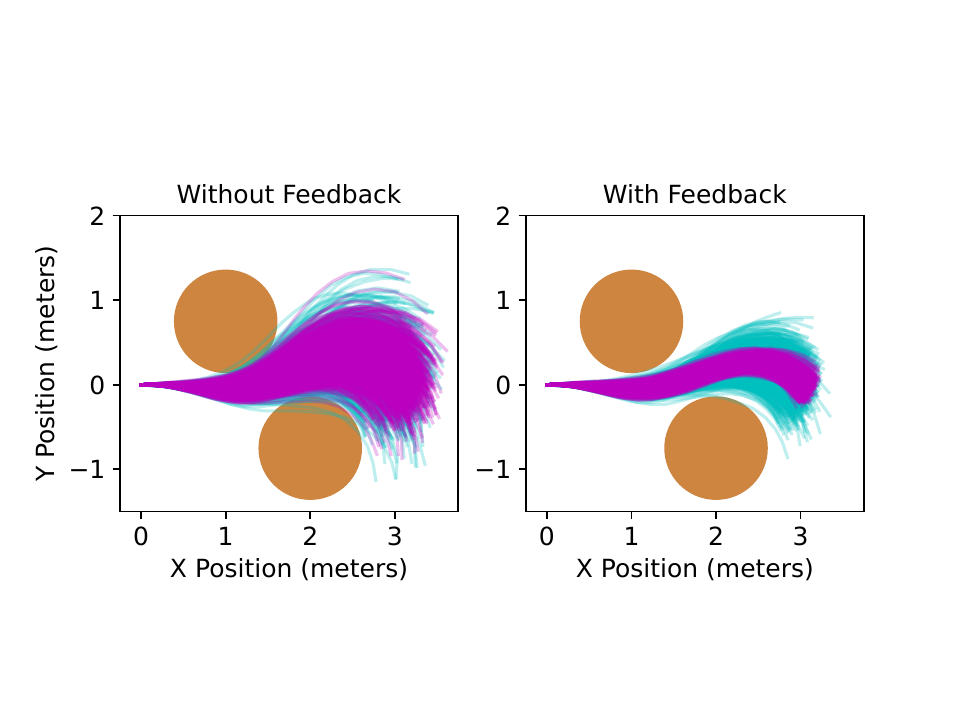}
    \caption{Samples from optimized policy distribution and stochastic dynamics in cyan. Samples from mean of the optimized policy distribution and stochastic dynamics in magenta. Obstacles in brown.}
    \label{fig:ugv_trajgen_samples}
\end{figure}

We simulate a stochastic bicycle model with acceleration and steering rate inputs. We denote ${\bx_t}=[ x_0, x_1, x_2, x_3, x_4]^T$ as the state vector, ${\bu_t}=[u_0, u_1]^T$ as the control vector, $l = 0.33$ as the wheel base, and $\boldsymbol{\Gamma} = diag(\left[0.001, 0.001, 0.1, 0.2, 0.001\right])$ as the covariance.
\begin{align}
  \bx_{t+1} &\sim p(\cdot | \bx_t, \bu_t) \triangleq \bx_t + \left(f(\bx_t, \bu_t) + \boldsymbol{\omega} \right) \Delta t \nonumber \\
  f(\bx_t, \bu_t) &= [x_3\cos(x_2), x_3\sin(x_2), x_3\tan(x_4) / l, u_0, u_1] \nonumber \\
  \vectg{\omega} &\sim \mathcal{N}(\cdot | \mathbf{0}, \boldsymbol{\Gamma})
\end{align}



A 20 timestep trajectory with $\Delta t=0.1$ sec is optimized, with an initial state of ${\bf x_I}=\left[0.0, 0.0, 0.0, 1.0, 0.0\right]^T$ and a goal state of ${\bf x_G}=\left[3.0, 0.0, 0.0, 1.0, 0.0\right]^T$. The steering angle is constrained to $(-0.4, 0.4)\ rad$, the steering rate to $(-1.0, 1.0)\ \frac{rad}{sec}$, and the acceleration to $(-1.0, 1.0)\ \frac{m}{s^2}$. We apply a quadratic cost on the final state of the trajectory: $J(\bX)=\bx_{N_T+1}^T\bQ_f\bx_{N_T+1}$ where $\vect{Q_f}=diag(\left[2.0, 2.0, 0.0, 0.0, 0.0\right])$. An obstacle constraint is applied with circular obstacles at $\left(1.0, 0.75\right)$ and $\left(2.0, -0.75\right)$. The initial prior distribution has a mean of all zeros and a variance of all ones. Parameters were as follows: $L=5$, $M=1024$, $\delta=0.05$, $\gamma=10$. Simulations were run on a laptop with an Intel Core i9-9880H CPU and a Nvidia GeForce RTX 2070 GPU.


We ran trajectory optimization for 500 iterations. Once completed, our method returns the hyper-parameters, $\widehat{\bnu}^*$, and the optimized PAC bounds, $\mathcal{J}^+_\alpha(\widehat{\bnu}^*)$ and $\mathcal{C}^+_\alpha(\widehat{\bnu}^*)$. It is guaranteed that policies sampled from $p(\bU|\bnu^*)$ will have an expected cost of $\leq\mathcal{J}^+_\alpha(\widehat{\bnu}^*)$ and an expected constraint (probability of collision) of $\leq\mathcal{C}^+_\alpha(\widehat{\bnu}^*)$ with a 95\% chance. To validate the PAC bounds, we compare them against Monte Carlo estimates of the expected cost, $\widehat{\mathcal J}(\widehat{\bnu}^*)$, and probability of collision, $\widehat{\mathcal C}(\widehat{\bnu}^*)$, at each iteration.


\begin{figure}[!t]
    \centering
    \includegraphics[trim={20 32 30 15},clip,width=1\columnwidth]{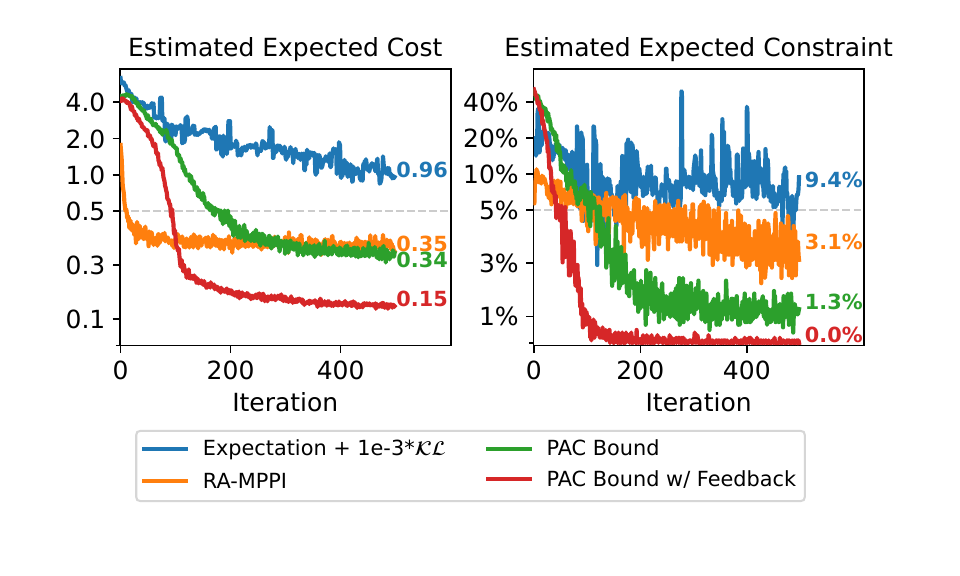}
    \caption{Comparison between our approach and other methods.}
    \label{fig:comparison}
\end{figure}

To highlight the necessity of properly considering stochasticity to generate accurate guarantees, we compared performance while optimizing the PAC bounds with and without considering stochastic dynamics (replace line 3 of Alg. \ref{alg:traj_model} and line 9 of Alg. \ref{alg:feedback} with $\bx_{t+1} = \bx_{t} + f(\bx_t, \bu_t)\Delta{t}$). Optimizing without stochastic dynamics caused $\widehat{\mathcal J}(\widehat{\bnu}^*)$ and $\widehat{\mathcal C}(\widehat{\bnu}^*)$ to be larger than the bounds, indicating that the guarantees were not accurate (Fig. \ref{fig:ugv_bounds}a,b). We also performed an ablation study in which we compared performance with and without feedback. On average, each iteration took 15.9 ms without feedback and 18.5 ms with feedback. Feedback enabled the optimizer to reduce the bounds to lower values (Fig. \ref{fig:ugv_bounds}e-h) and resulted in a tighter distribution in state-space (Fig \ref{fig:ugv_trajgen_samples}).


We compare against optimizing an empirical approximation of the expected costs/constraints, regulated by the $\mathcal{KL}$-divergence to prior policies, to demonstrate that optimizing the PAC bounds encourages more robust polices. We also compare against RA-MPPI \cite{yin2023risk}, a state-of-the-art risk-aware sampling based NMPC method. We used the following parameters: $\Sigma_\epsilon=0.01$, $M=1024$, $N=300$, $\eta=0$, $\alpha=0.3$, $A=10$, $B=1$, $C_u=0.5$ for the cost, and $C_u=0.05$ for the constraint. Our approach converged to the lowest estimated expected cost and probability of collision (Fig \ref{fig:comparison}).

\subsection{NMPC Experiment}

\begin{figure}[!t]%
    \centering
    \includegraphics[trim={15 20 45 55},clip,width=.85\columnwidth]{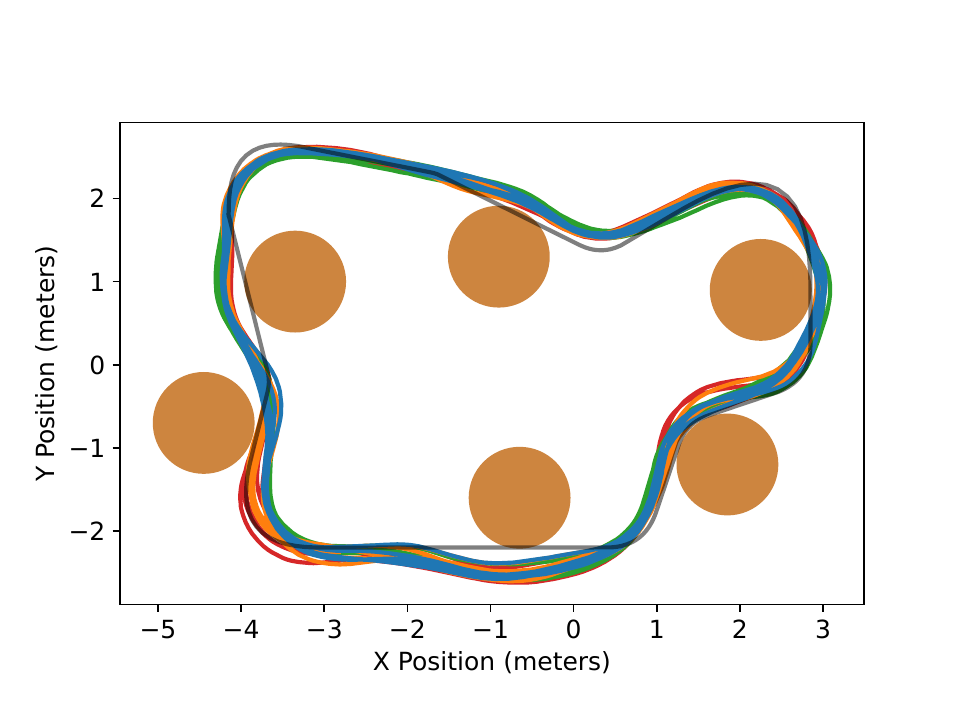}
    \caption{Paths taken while following route (black) and avoiding obstacles (brown) in simulation. Blue/green paths use stochastic bounds, red/orange paths do not. Blue/orange paths use feedback, red/green paths do not.}
    \label{fig:ugv_nmpc_route}
\end{figure}

The system follows a receding horizon state generated on a looping path. The path runs along a set of circular obstacles. We optimize a 12 timestep trajectory with $\Delta t=0.1$ second at a replanning period of $H=0.2$ sec. We apply a quadratic cost on the final state of the trajectory: $J(\bX)=\bx_{N_T+1}^T\bQ_f\bx_{N_T+1}$ where $\vect{Q_f}=diag(\left[1.0, 1.0, 0.1, 0.1, 0.0\right])$. Parameters were as follows: $L=5$, $M=1024$, $\delta=0.05$, $\gamma=10$. To allow for a higher control rate, the executed control trajectory is interpolated from 10 Hz to 50 Hz with the assumption that our PAC bounds will still hold. We perform an ablation study in which we compare performance with and without feedback, as well as optimizing the PAC bounds with and without stochastic dynamics.

PAC-NMPC was successfully able to control the system in real time and followed the path while avoiding nearby obstacles. In Figure \ref{fig:ugv_nmpc_route}, we show the route taken by the system during five loops of the path.

\begin{figure}[!t]%
  \centering
    \includegraphics[trim={18 0 45 0}, clip, width=0.9\columnwidth]{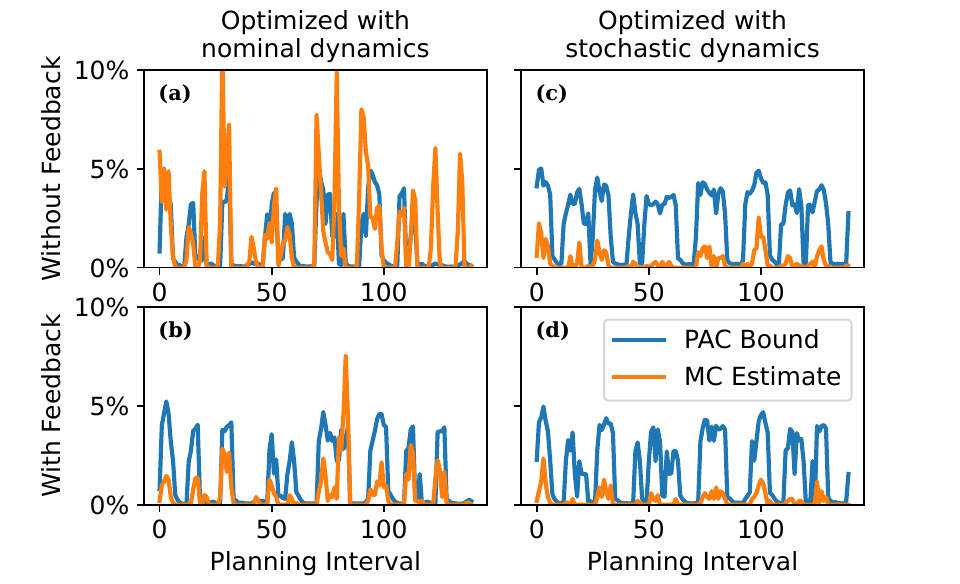}
    \caption{Optimized PAC Bound compared to Monte Carlo estimates for probability of constraint violation at each planning interval during first laps. Variations in the PAC Bound and MC estimate over time are expected behavior due to the robot moving closer/farther from obstacles along the route.}
  \label{fig:ugv_nmpc_bounds}
\end{figure}

During each planning interval of $H=0.2$ sec, the controller achieved an average of 54 iterations of PAC stochastic trajectory optimization (approximately 3.7ms per iteration). The controller consistently produced PAC bounds $\leq$ 5\% probability of collision, which accurately bounded the Monte Carlo estimates when optimizing with stochastic dynamics. The bound was exceeded only once out of 482 planning intervals. Thus, the estimates were bounded accurately in around 99.8\% of planning intervals, well within the 95\% confidence bound (Fig \ref{fig:ugv_nmpc_bounds}d).


We demonstrate a quantitative difference in the observance of the PAC Bounds in Figure \ref{fig:ugv_nmpc_bounds}. When optimizing the bounds with the nominal dynamics, the probability of collision estimates were only bounded in 93.9\% of planning intervals (Fig \ref{fig:ugv_nmpc_bounds}b). This decrease in bound accuracy is even more apparent when running the controller without feedback to compensate for unmodeled noise, resulting in only 64.7\% of planning intervals having bounded probability of collision estimates (Fig \ref{fig:ugv_nmpc_bounds}a).

\section{Hardware Experiments}

\subsection{Rally Car}

\subsubsection{Hardware}


To evaluate our method on physical hardware, we ran PAC-NMPC on a 1/10th-scale Traxxas Rally Car platform. The algorithm runs on a Nvidia Jetson Orin mounted on the bottom platform. The control interface is a variable electronic speed controller (VESC), which executes commands and provides servo state information. The position and orientation is tracked with an OptiTrack system.

\subsubsection{Setup}

The route, virtual obstacle placements, and TVLQR costs are identical to those in the simulation experiments. We optimize a 12 timestep trajectory with $\Delta t=0.1$ sec at a replanning period of $H=0.2$ sec. We apply a quadratic cost to the final state of the trajectory: $J(\bX)=\bx_{N_T+1}\bQ_f\bx_{N_T+1}$ where $\vect{Q_f}=diag(\left[1.0, 1.0, 0.4, 0.1, 0.0\right])$. Parameters were $L=2$, $M=1024$, $\delta=0.05$, $\gamma=10$.

\begin{figure}[!t]%
  \includegraphics[trim={25 10 50 0}, clip, width=0.9\columnwidth]{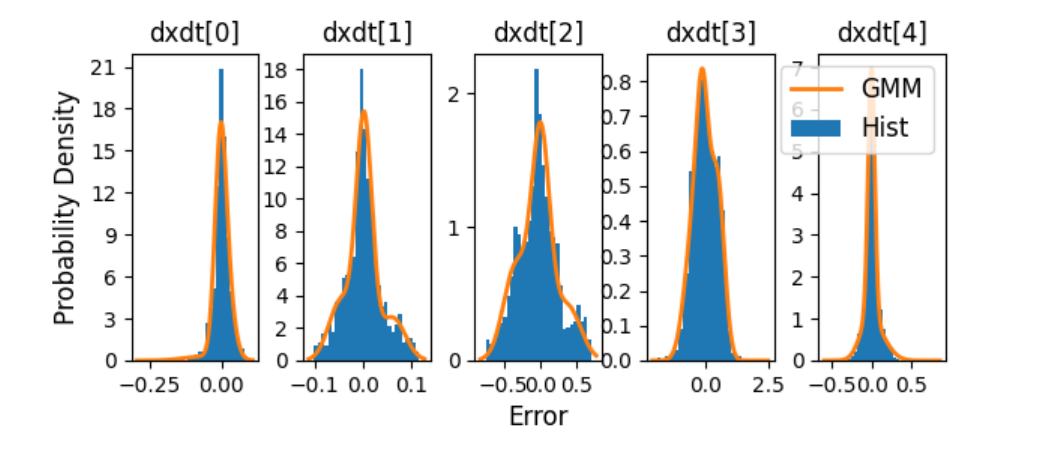}
  \caption{Rallycar dynamics noise model}
  \label{fig:ugv_dynamics_noise}
\end{figure}

We model the rally car dynamics as a stochastic bicycle model where noise is modeled as a 3 component Gaussian Mixture Model. This mixture model was fit from 2500 samples of data collected with the car (Fig. \ref{fig:ugv_dynamics_noise}). We perform the same ablation study as done in the simulation experiments.

\subsubsection{Results}

\begin{figure}[!t]%
    \centering
      \includegraphics[trim={15 5 45 44},clip,width=0.85\columnwidth]{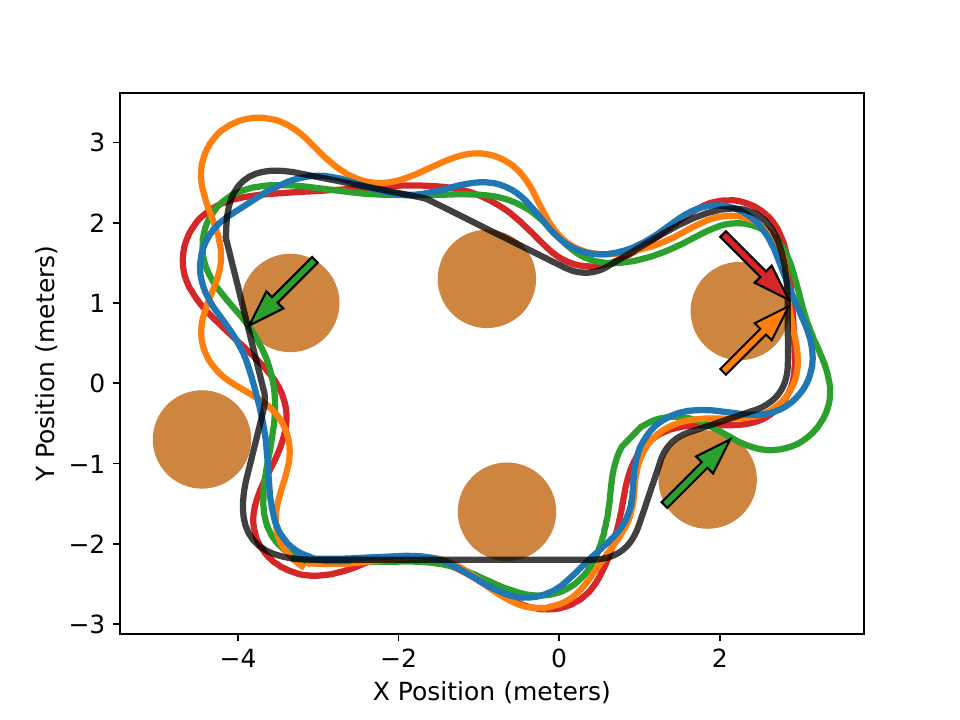}
      \caption{Path taken while following route (black) and avoiding obstacles (brown) with rally car hardware. Blue/green paths use stochastic bounds, red/orange paths do not. Blue/orange paths use feedback, red/green paths do not. Collisions are annotated by arrows.}
    \label{fig:ugv_hardware_route}
\end{figure}

PAC-NMPC ran onboard the rally car in real-time and followed the path while avoiding nearby obstacles. In Figure \ref{fig:ugv_hardware_route}, we display the path taken by the car with collisions annotated by arrows. During each planning interval of $H=0.2$ sec, the controller achieved an average of 10 iterations of trajectory distribution optimization with approximately 20ms per iteration.

\begin{figure}[!t]%
    \centering
      \includegraphics[trim={18 0 40 0}, clip, width=.9\columnwidth]{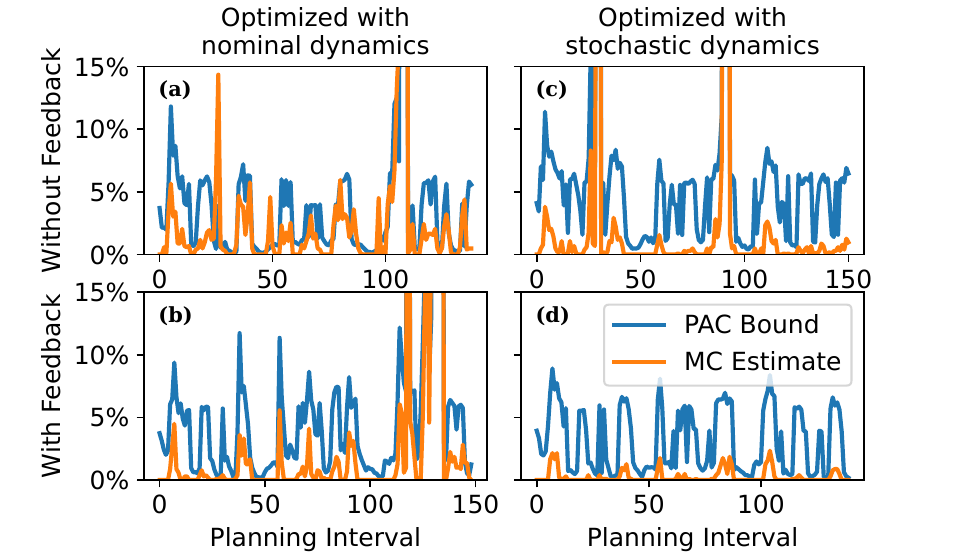}
      \caption{Optimized PAC Bound compared to Monte Carlo estimates for probability of constraint violation at each planning interval.}
    \label{fig:ugv_hardware_bound}
\end{figure}

The controller consistently produced PAC bounds $\leq$10\% probability of collision, which accurately bounded the Monte Carlo estimates of the probability of collision at all planning intervals (Fig. \ref{fig:ugv_hardware_bound}d). In this context, Monte Carlo estimates refer to sampled trajectories using the simulated stochastic dynamics model. When optimizing with stochastic dynamics and without feedback, the car collided with virtual obstacles twice, likely due to unmodeled dynamics not captured in the Gaussian mixture model (Fig. \ref{fig:ugv_hardware_bound}c). When optimizing the bound with nominal dynamics, the car collided with virtual obstacles, with and without feedback (Fig. \ref{fig:ugv_hardware_bound}a,b).

\subsection{Fixed-Wing UAV}

\subsubsection{Hardware}
To demonstrate that our approach can extend to more complex, high-dimensional systems, we use PAC-NMPC to control a 24" wingspan Edge 540 fixed-wing UAV. It has a controllable propeller, rudder, elevator, and kinematically linked ailerons. The position and orientation are tracked with an OptiTrack system.

\subsubsection{Setup}
\begin{figure}[!t]%
    \centering
      \includegraphics[trim={0 0 30 20},clip,width=0.8\columnwidth]{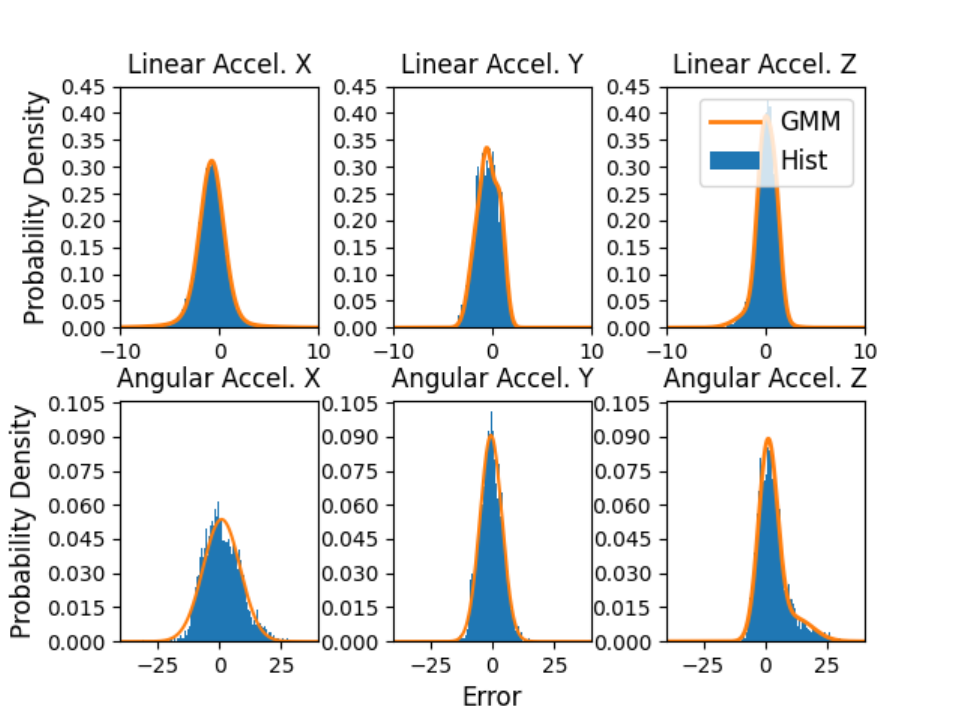}
      \caption{Edge540 Acceleration Noise Model}
    \label{fig:edge540_noise}
\end{figure}

We utilize a quaternion formulation of the fixed-wing described in \cite{basescu2020direct} with RK2 integration. This system has a 17-dimensional state space and a 4-dimensional control space. The state of the system is $\vect{x} = \left[\vect{r}, \vect{q}, \vectg{\delta}, \vect{v}, \vectg{\omega}, p\right]$ where $\vect{r} \in \mathbb{R}^3$ is the position, $\vect{q} \in \mathbb{Q}$ is the orientation, $\vectg{\delta} \in \mathbb{R}^3$ are control surface deflections of the ailerons, elevator, and rudder respectively, $\vect{v} \in \mathbb{R}^3$ is the linear velocity, $\vectg{\omega} \in \mathbb{R}^3$ is the angular velocity, and $p$ is the propeller speed. The control signal is the rate of change of the control surface deflections and of the propeller speed: $\vect{u} = \left[u_a, u_e, u_r, u_p \right]$. We place noise represented by a 3 component Gaussian Mixture Model over the body frame accelerations, which was fit from 2000 samples of data collected with the fixed-wing (Fig. \ref{fig:edge540_noise}).
We optimize a 12 timestep trajectory with $\Delta t = 0.1$ sec at a replanning period of $H = 0.2$. We apply a quadratic cost of the form $J(\bX) = \vect{x_{N_T+1}}^T\vect{Q_f}\vect{x_{N_T+1}} + \sum_{t=1}^{N_T} \left(\vect{x_t}^T\vect{Q}\vect{x_t} + \vect{u_t}^T\vect{R}\vect{u_t}\right)$ and an obstacle constraint. Parameters were as follows: $L=1$, $M=1024$, $\delta=0.05$, and $\gamma=10$. We use feedback and use stochastic dynamics to optimize the PAC bounds. 



\begin{figure}[t]%
    \centering
      \includegraphics[trim={0 0 0 0},clip,width=0.85\columnwidth]{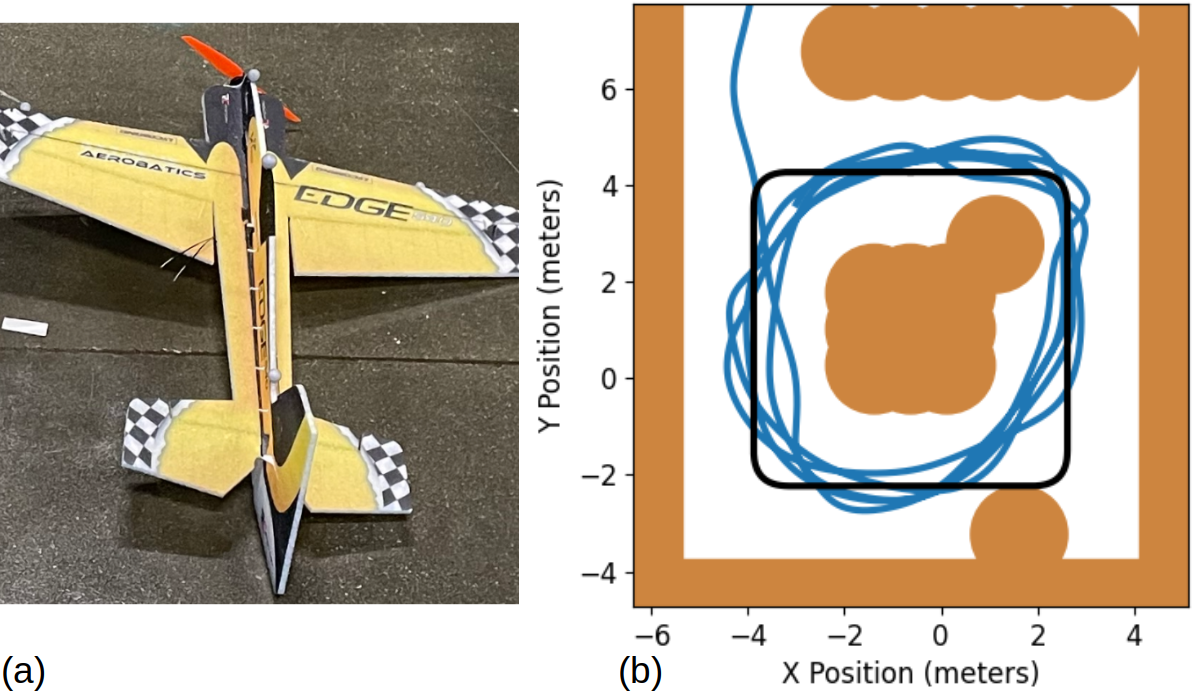}
      \caption{a) Edge540 b) Path (blue) taken by the fixed-wing while following route (black) and avoiding obstacles (brown).}
    \label{fig:uav_nmpc_route}
\end{figure}

\begin{figure}[t]%
    \centering
      \includegraphics[trim={0 0 40 20}, clip, width=1\columnwidth]{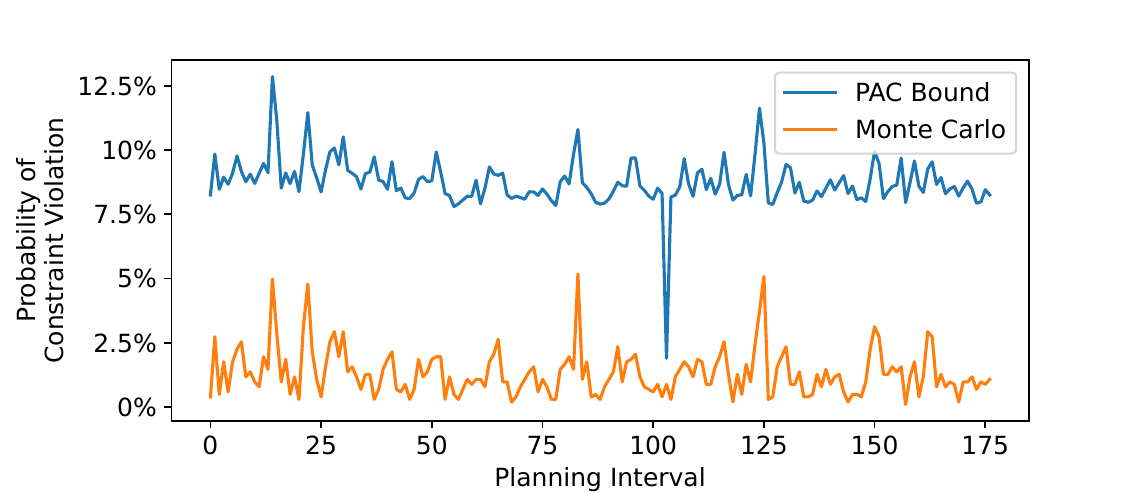}
      \caption{Optimized PAC Bound compared to Monte Carlo estimates for probability of constraint violation.}
    \label{fig:uav_nmpc_bound}
\end{figure}
To enable real-time performance, we set the number of prior policies, $L$, to one and compute the linearized dynamics for the feedback policy using finite difference on the mean trajectory of the prior. The controller was run on a desktop computer with an AMD Ryzen 7 5800x and a NVIDIA GeForce RTX 3080.
\subsubsection{Results}
PAC-NMPC was successfully able to control the fixed-wing in real-time, following a square path while avoiding nearby obstacles. In Figure \ref{fig:uav_nmpc_route}, we display the route taken by the system during five loops of the path. During each planning interval of $H = 0.2$ sec, the controller achieved an average of 22 iterations of trajectory distribution optimization with around 8.5 ms per iteration. The controller consistently produced PAC bounds of around 10\% probability of collision, which accurately bounded the Monte Carlo estimates (Fig. \ref{fig:uav_nmpc_bound}).
\section{Discussion}

In this work, we presented a novel SNMPC method capable of propagating uncertainty through arbitrary nonlinear dynamic systems and of providing statistical guarantees on expected cost and constraint violations. We demonstrated real time performance both in simulation and on-board physical hardware. Further, we show that the algorithm is capable of scaling to more complex systems, like fixed-wing UAVs. Since this algorithm can be used with ``black-box'' sampling of dynamics (assuming they are continuously differentiable), costs, constraints, and noise, future work can investigate its use with learned dynamics and perception-informed costs. Future work could also explore the extension of this approach to more complex control trajectory distributions.

\bibliographystyle{IEEEtran}
\bibliography{references}

\begin{thebibliography}{10}
\providecommand{\url}[1]{#1}
\csname url@samestyle\endcsname
\providecommand{\newblock}{\relax}
\providecommand{\bibinfo}[2]{#2}
\providecommand{\BIBentrySTDinterwordspacing}{\spaceskip=0pt\relax}
\providecommand{\BIBentryALTinterwordstretchfactor}{4}
\providecommand{\BIBentryALTinterwordspacing}{\spaceskip=\fontdimen2\font plus
\BIBentryALTinterwordstretchfactor\fontdimen3\font minus \fontdimen4\font\relax}
\providecommand{\BIBforeignlanguage}[2]{{%
\expandafter\ifx\csname l@#1\endcsname\relax
\typeout{** WARNING: IEEEtran.bst: No hyphenation pattern has been}%
\typeout{** loaded for the language `#1'. Using the pattern for}%
\typeout{** the default language instead.}%
\else
\language=\csname l@#1\endcsname
\fi
#2}}
\providecommand{\BIBdecl}{\relax}
\BIBdecl

\bibitem{falanga2018pampc}
D.~Falanga, P.~Foehn, P.~Lu, and D.~Scaramuzza, ``Pampc: Perception-aware model predictive control for quadrotors,'' in \emph{2018 IEEE/RSJ International Conference on Intelligent Robots and Systems (IROS)}.\hskip 1em plus 0.5em minus 0.4em\relax IEEE, 2018, pp. 1--8.

\bibitem{basescu2020direct}
M.~Basescu and J.~Moore, ``Direct nmpc for post-stall motion planning with fixed-wing uavs,'' in \emph{2020 IEEE International Conference on Robotics and Automation (ICRA)}.\hskip 1em plus 0.5em minus 0.4em\relax IEEE, 2020, pp. 9592--9598.

\bibitem{ding2019real}
Y.~Ding, A.~Pandala, and H.-W. Park, ``Real-time model predictive control for versatile dynamic motions in quadrupedal robots,'' in \emph{2019 International Conference on Robotics and Automation (ICRA)}.\hskip 1em plus 0.5em minus 0.4em\relax IEEE, 2019, pp. 8484--8490.

\bibitem{grandia2019feedback}
R.~Grandia, F.~Farshidian, R.~Ranftl, and M.~Hutter, ``Feedback mpc for torque-controlled legged robots,'' in \emph{2019 IEEE/RSJ International Conference on Intelligent Robots and Systems (IROS)}.\hskip 1em plus 0.5em minus 0.4em\relax IEEE, 2019, pp. 4730--4737.

\bibitem{paulson2019efficient}
J.~A. Paulson and A.~Mesbah, ``An efficient method for stochastic optimal control with joint chance constraints for nonlinear systems,'' \emph{International Journal of Robust and Nonlinear Control}, vol.~29, no.~15, pp. 5017--5037, 2019.

\bibitem{manchester2017dirtrel}
Z.~Manchester and S.~Kuindersma, ``Dirtrel: Robust trajectory optimization with ellipsoidal disturbances and lqr feedback.'' in \emph{Robotics: Science and Systems}.\hskip 1em plus 0.5em minus 0.4em\relax Cambridge, MA, USA, 2017.

\bibitem{singh2017robust}
S.~Singh, A.~Majumdar, J.-J. Slotine, and M.~Pavone, ``Robust online motion planning via contraction theory and convex optimization,'' in \emph{2017 IEEE International Conference on Robotics and Automation (ICRA)}.\hskip 1em plus 0.5em minus 0.4em\relax IEEE, 2017, pp. 5883--5890.

\bibitem{ozaki2020tube}
N.~Ozaki, S.~Campagnola, and R.~Funase, ``Tube stochastic optimal control for nonlinear constrained trajectory optimization problems,'' \emph{Journal of Guidance, Control, and Dynamics}, 2020.

\bibitem{yin2022trajectory}
J.~Yin, Z.~Zhang, E.~Theodorou, and P.~Tsiotras, ``Trajectory distribution control for model predictive path integral control using covariance steering,'' in \emph{2022 International Conference on Robotics and Automation (ICRA)}.\hskip 1em plus 0.5em minus 0.4em\relax IEEE, 2022, pp. 1478--1484.

\bibitem{kobilarov2015sample}
M.~Kobilarov, ``Sample complexity bounds for iterative stochastic policy optimization,'' \emph{Advances in Neural Information Processing Systems}, vol.~28, 2015.

\bibitem{sheckells2020probably}
M.~Sheckells, ``Probably approximately correct robust policy search with applications to mobile robotics,'' Ph.D. dissertation, Johns Hopkins University, 2020.

\bibitem{villanueva2017robust}
M.~E. Villanueva, R.~Quirynen, M.~Diehl, B.~Chachuat, and B.~Houska, ``Robust mpc via min--max differential inequalities,'' \emph{Automatica}, vol.~77, pp. 311--321, 2017.

\bibitem{garimella2018robust}
G.~Garimella, M.~Sheckells, J.~L. Moore, and M.~Kobilarov, ``Robust obstacle avoidance using tube nmpc.'' in \emph{Robotics: Science and Systems}, 2018.

\bibitem{lucia2014multi}
S.~Lucia, R.~Paulen, and S.~Engell, ``Multi-stage nonlinear model predictive control with verified robust constraint satisfaction,'' in \emph{53rd IEEE Conference on Decision and Control}.\hskip 1em plus 0.5em minus 0.4em\relax IEEE, 2014, pp. 2816--2821.

\bibitem{mayne2011tube}
D.~Q. Mayne, E.~C. Kerrigan, E.~Van~Wyk, and P.~Falugi, ``Tube-based robust nonlinear model predictive control,'' \emph{International journal of robust and nonlinear control}, vol.~21, no.~11, pp. 1341--1353, 2011.

\bibitem{marruedo2002input}
D.~L. Marruedo, T.~Alamo, and E.~F. Camacho, ``Input-to-state stable mpc for constrained discrete-time nonlinear systems with bounded additive uncertainties,'' in \emph{Proceedings of the 41st IEEE Conference on Decision and Control, 2002.}, vol.~4.\hskip 1em plus 0.5em minus 0.4em\relax IEEE, 2002, pp. 4619--4624.

\bibitem{gao2014tube}
Y.~Gao, A.~Gray, H.~E. Tseng, and F.~Borrelli, ``A tube-based robust nonlinear predictive control approach to semiautonomous ground vehicles,'' \emph{Vehicle System Dynamics}, vol.~52, no.~6, pp. 802--823, 2014.

\bibitem{bayer2013discrete}
F.~Bayer, M.~B{\"u}rger, and F.~Allg{\"o}wer, ``Discrete-time incremental iss: A framework for robust nmpc,'' in \emph{2013 European Control Conference (ECC)}.\hskip 1em plus 0.5em minus 0.4em\relax IEEE, 2013, pp. 2068--2073.

\bibitem{chou2021model}
G.~Chou, N.~Ozay, and D.~Berenson, ``Model error propagation via learned contraction metrics for safe feedback motion planning of unknown systems,'' in \emph{2021 60th IEEE Conference on Decision and Control (CDC)}.\hskip 1em plus 0.5em minus 0.4em\relax IEEE, 2021, pp. 3576--3583.

\bibitem{kohler2020computationally}
J.~K{\"o}hler, R.~Soloperto, M.~A. M{\"u}ller, and F.~Allg{\"o}wer, ``A computationally efficient robust model predictive control framework for uncertain nonlinear systems,'' \emph{IEEE Transactions on Automatic Control}, vol.~66, no.~2, pp. 794--801, 2020.

\bibitem{lew2021sampling}
T.~Lew and M.~Pavone, ``Sampling-based reachability analysis: A random set theory approach with adversarial sampling,'' in \emph{Conference on robot learning}.\hskip 1em plus 0.5em minus 0.4em\relax PMLR, 2021, pp. 2055--2070.

\bibitem{wu2023robust}
A.~Wu, T.~Lew, K.~Solovey, E.~Schmerling, and M.~Pavone, ``Robust-rrt: Probabilistically-complete motion planning for uncertain nonlinear systems,'' in \emph{Robotics Research}.\hskip 1em plus 0.5em minus 0.4em\relax Springer, 2023, pp. 538--554.

\bibitem{todorov2005generalized}
E.~Todorov and W.~Li, ``A generalized iterative lqg method for locally-optimal feedback control of constrained nonlinear stochastic systems,'' in \emph{Proceedings of the 2005, American Control Conference, 2005.}\hskip 1em plus 0.5em minus 0.4em\relax IEEE, 2005, pp. 300--306.

\bibitem{todorov2009iterative}
E.~Todorov and Y.~Tassa, ``Iterative local dynamic programming,'' in \emph{2009 IEEE Symposium on Adaptive Dynamic Programming and Reinforcement Learning}.\hskip 1em plus 0.5em minus 0.4em\relax IEEE, 2009, pp. 90--95.

\bibitem{theodorou2010stochastic}
E.~Theodorou, Y.~Tassa, and E.~Todorov, ``Stochastic differential dynamic programming,'' in \emph{Proceedings of the 2010 American Control Conference}.\hskip 1em plus 0.5em minus 0.4em\relax IEEE, 2010, pp. 1125--1132.

\bibitem{howell2021direct}
T.~A. Howell, C.~Fu, and Z.~Manchester, ``Direct policy optimization using deterministic sampling and collocation,'' \emph{IEEE Robotics and Automation Letters}, vol.~6, no.~3, pp. 5324--5331, 2021.

\bibitem{williams2016aggressive}
G.~Williams, P.~Drews, B.~Goldfain, J.~M. Rehg, and E.~A. Theodorou, ``Aggressive driving with model predictive path integral control,'' in \emph{2016 IEEE International Conference on Robotics and Automation (ICRA)}.\hskip 1em plus 0.5em minus 0.4em\relax IEEE, 2016, pp. 1433--1440.

\bibitem{williams2018robust}
G.~Williams, B.~Goldfain, P.~Drews, K.~Saigol, J.~M. Rehg, and E.~A. Theodorou, ``Robust sampling based model predictive control with sparse objective information.'' in \emph{Robotics: Science and Systems}, 2018.

\bibitem{gandhi2021robust}
M.~S. Gandhi, B.~Vlahov, J.~Gibson, G.~Williams, and E.~A. Theodorou, ``Robust model predictive path integral control: Analysis and performance guarantees,'' \emph{IEEE Robotics and Automation Letters}, vol.~6, no.~2, pp. 1423--1430, 2021.

\bibitem{yin2023risk}
J.~Yin, Z.~Zhang, and P.~Tsiotras, ``Risk-aware model predictive path integral control using conditional value-at-risk,'' in \emph{2023 IEEE International Conference on Robotics and Automation (ICRA)}.\hskip 1em plus 0.5em minus 0.4em\relax IEEE, 2023, pp. 7937--7943.

\bibitem{deisenroth2013survey}
M.~P. Deisenroth, G.~Neumann, J.~Peters \emph{et~al.}, ``A survey on policy search for robotics,'' \emph{Foundations and Trends{\textregistered} in Robotics}, vol.~2, no. 1--2, pp. 1--142, 2013.

\bibitem{majumdar2021pac}
A.~Majumdar, A.~Farid, and A.~Sonar, ``Pac-bayes control: learning policies that provably generalize to novel environments,'' \emph{The International Journal of Robotics Research}, vol.~40, no. 2-3, pp. 574--593, 2021.

\bibitem{veer2021probably}
S.~Veer and A.~Majumdar, ``Probably approximately correct vision-based planning using motion primitives,'' in \emph{Conference on Robot Learning}.\hskip 1em plus 0.5em minus 0.4em\relax PMLR, 2021, pp. 1001--1014.

\bibitem{peters2010relative}
J.~Peters, K.~Mulling, and Y.~Altun, ``Relative entropy policy search,'' in \emph{Proceedings of the AAAI Conference on Artificial Intelligence}, vol.~24, no.~1, 2010, pp. 1607--1612.

\bibitem{catoni2012challenging}
O.~Catoni, ``Challenging the empirical mean and empirical variance: a deviation study,'' in \emph{Annales de l'IHP Probabilit{\'e}s et statistiques}, vol.~48, no.~4, 2012, pp. 1148--1185.

\bibitem{fei2014parallel}
Y.~Fei, G.~Rong, B.~Wang, and W.~Wang, ``Parallel l-bfgs-b algorithm on gpu,'' \emph{Computers \& graphics}, vol.~40, pp. 1--9, 2014.

\bibitem{underactuated}
\BIBentryALTinterwordspacing
R.~Tedrake, \emph{Underactuated Robotics}, 2022. [Online]. Available: \url{http://underactuated.mit.edu}
\BIBentrySTDinterwordspacing

\end{thebibliography}

\end{document}